%% file: main.tex
\documentclass[]{beingbeyond}
\usepackage{enumitem}
\usepackage[toc,page,header]{appendix}

\usepackage[utf8]{inputenc} 
\usepackage[T1]{fontenc}    
\usepackage{hyperref}       
\usepackage{url}            
\usepackage{array}          
\usepackage{booktabs}       
\usepackage{amsfonts}       
\usepackage{nicefrac}       
\usepackage{microtype}      
\usepackage{xcolor}         
\usepackage{xspace}
\usepackage{bm}
\usepackage{bbm}
\usepackage{bbding}
\usepackage{tabularx}
\usepackage{textcomp}
\usepackage{amssymb}
\usepackage{enumitem}
\usepackage{amsmath}
\usepackage{mathtools}
\usepackage{amsthm}
\usepackage{multirow}
\usepackage{makecell}
\usepackage{color}
\usepackage{colortbl}
\usepackage{adjustbox}
\usepackage{caption}
\usepackage{graphicx}
\usepackage{wrapfig}
\usepackage{array}
\usepackage{multicol}

\definecolor{myyellow}{RGB}{255,192,0}
\definecolor{mygreen}{RGB}{107,170,64}
\definecolor{mywrite}{RGB}{255,227,132}

\title{Spatial-Aware VLA Pretraining through Visual-Physical Alignment from Human Videos}

\author{{\bfseries 
Yicheng Feng$^{1,3}$ \quad
Wanpeng Zhang$^{1,3}$ \quad
Ye Wang$^{2,3}$ \quad
Hao Luo$^{1,3}$ \\
Haoqi Yuan$^{1,3}$ \quad
Sipeng Zheng$^{3}$ \quad
Zongqing Lu$^{1,3,\dagger}$
}}

\affiliation{{$^{1}$Peking University \quad $^{2}$Renmin University of China \quad $^{3}$BeingBeyond}}

\webpage{\url{https://beingbeyond.github.io/VIPA-VLA}}

\input{section/00_abstract}

\checkdata[Date]{December 15, 2025}

\definecolor{BlockC}{gray}{0.98}  
\definecolor{BlockA}{RGB}{191,211,230}
\definecolor{BlockB}{RGB}{199,233,192}

\newcommand{\DataName}{\textbf{\texttt{Hand3D}}\xspace}
\newcommand{\DataInstName}{\textbf{\texttt{Hand3D-visual}}\xspace}
\newcommand{\DataTransName}{\textbf{\texttt{Hand3D-action}}\xspace}
\newcommand{\DataTestName}{\textbf{\texttt{Hand3D-test}}\xspace}
\newcommand{\ModelName}{\textbf{\texttt{VIPA-VLA}}\xspace}

\begingroup
\footnotetext[2]{Correspondence to Zongqing Lu $<$lu@beingbeyond.com$>$.}
\endgroup

\begin{document}

\maketitle

\input{section/01_intro}
\input{section/02_relatedwork}
\input{section/03-1_overview}
\input{section/05_experiment}
\input{section/06_conclusion}

\clearpage

\bibliographystyle{unsrt}
\bibliography{ref}





\clearpage

\end{document}

%% file: section/00_abstract.tex
\abstract{
Vision-Language-Action (VLA) models provide a promising paradigm for robot learning by integrating visual perception with language-guided policy learning. However, most existing approaches rely on 2D visual inputs to perform actions in 3D physical environments, creating a significant gap between perception and action grounding. To bridge this gap, we propose a \textbf{Spatial-Aware VLA Pretraining} paradigm that performs explicit alignment between visual space and physical space during pretraining, enabling models to acquire 3D spatial understanding before robot policy learning. Starting from pretrained vision-language models, we leverage large-scale human demonstration videos to extract 3D visual and 3D action annotations, forming a new source of supervision that aligns 2D visual observations with 3D spatial reasoning. We instantiate this paradigm with \ModelName, a dual-encoder architecture that incorporates a 3D visual encoder to augment semantic visual representations with 3D-aware features.
When adapted to downstream robot tasks, \ModelName achieves significantly improved grounding between 2D vision and 3D action, resulting in more robust and generalizable robotic policies.
}

%% file: section/01_intro.tex
\section{Introduction}
\label{sec:intro}

\begin{figure*}[t]
    \centering
    \includegraphics[width=0.85\linewidth]{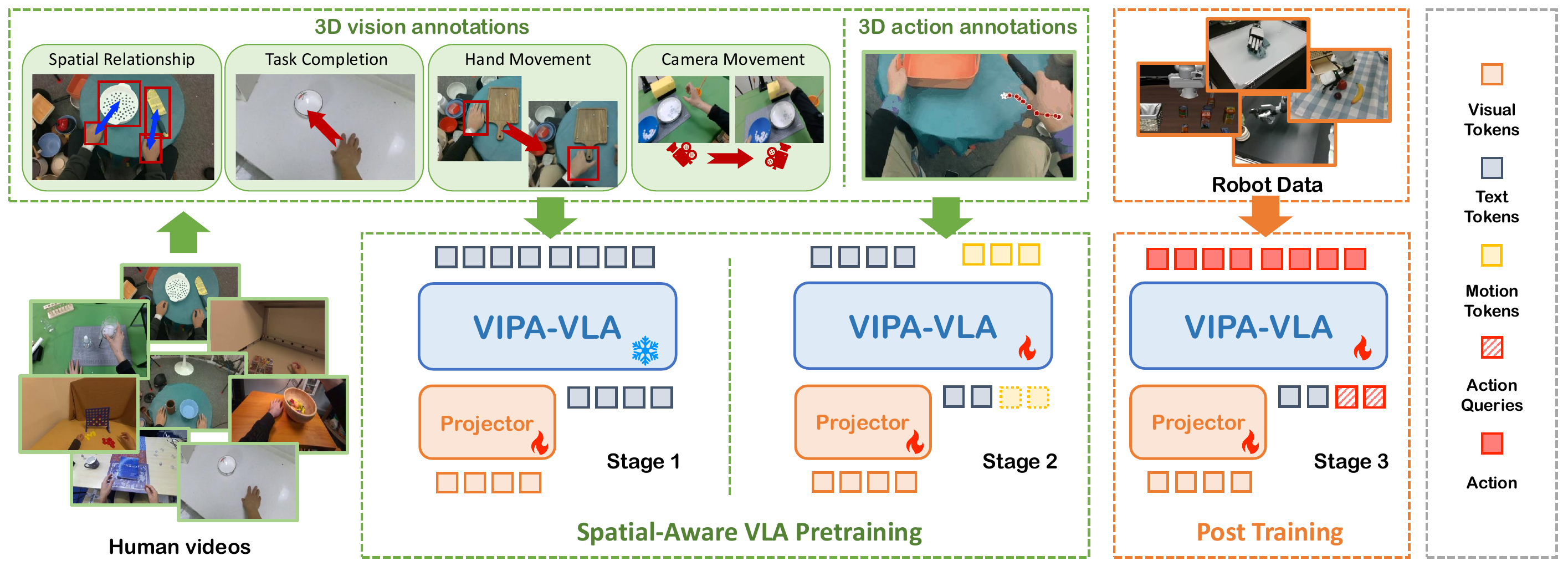}
    \caption{
    \textbf{Overview of the proposed \emph{Visual-Physical Alignment} framework.} We start from diverse human demonstration videos to extract 3D visual annotations and 3D action annotations, capturing how humans interact with the physical world with corresponding visual observations. These annotations enable the two-stage \textbf{Spatial-Aware VLA Pretraining}, which teaches VLA models to ground 2D visual inputs in 3D spatial understanding: (1) 3D-Visual Pretraining: starting from a VLM backbone, human demonstration videos with 3D visual annotations are used to align 2D visual features with 3D spatial representations via a dual-encoder fusion module. (2) 3D-Action Pretraining: human hand trajectories provide 3D motion supervision, enabling the model to learn physically grounded action priors. Then in the third stage, the pretrained model, \ModelName, is adapted to robot manipulation tasks, resulting in robust and generalizable policies in simulation and real-world settings.}
    \vspace{-3mm}
    \label{fig:intro}
\end{figure*}

The rapid progress of large-scale vision-language models (VLMs) has showcased remarkable capabilities in learning joint representations across modalities\citep{achiam2023gpt,team2023gemini,bai2025qwen2,zhu2025internvl3,Li2023BLIP2BL}. This progress has opened up new opportunities for robot policy learning, where VLMs offer a solid foundation for understanding both visual observations and language instructions, thereby guiding robot interaction with the physical world. Building on this foundation, the vision-language-action (VLA) paradigm has recently demonstrated the potential to develop generalist robot policies across a wide range of tasks\citep{black2410pi0,bjorck2025gr00t,brohan2022rt1,brohan2023rt2,kim2024openvla}.

Nevertheless, current VLA models typically rely on 2D visual inputs to perceive the world while performing actions in a 3D physical environment, leaving a substantial gap between visual perception and embodied action. This weak correspondence limits their ability to ground actions in physical space. For effective policy learning, an agent must not only interpret pixels but also understand how these visual cues map to 3D geometry and how physical actions interact with the surrounding environment. While humans can infer 3D space from 2D visual signals, existing VLA models largely overlook this aspect, resulting in poor spatial grounding and limited generalization.

To bridge this gap, we introduce a new \textbf{Spatial-Aware VLA Pretraining} paradigm that enables models to acquire 3D spatial awareness before learning robotic policies. Starting from pretrained VLMs, we leverage large-scale human demonstration videos as a rich source of supervision, where implicit correspondences between 2D visual observations and 3D physical actions naturally exist. 
Compared to robot data, human demonstrations are easier to obtain across diverse environments and naturally provide rich evidence of how actions are carried out in the physical world under diverse visual contexts.
By extracting 3D cues such as hand-object relationships and motion trajectories from these videos, we construct pretraining tasks that teach the model to align 2D vision with 3D spatial understanding—forming what we term \emph{visual-physical alignment}. This spatially grounded pretraining provides a strong foundation for subsequent VLA post-training, allowing the model to learn and generalize more effectively in robot manipulation tasks.

To instantiate this paradigm, we present \ModelName (\textbf{Vi}sual-\textbf{P}hysical-\textbf{A}lignment-\textbf{VLA}), a dual-encoder architecture that augments semantic visual representations with explicit 3D spatial features. Alignment between vision, language, and 3D action is achieved through Spatial-Aware VLA Pretraining on our dataset \DataName, which is constructed from diverse human manipulation recordings with annotated hand poses. From these videos, we derive two forms of spatial supervision: \textbf{3D visual annotations}, providing coarse-grained spatial grounding between visual inputs and physical configurations, and \textbf{3D action annotations}, offering fine-grained motion supervision through 3D trajectories that encode human manipulation dynamics. The pretrained model is then adapted to robotic tasks via post-training, transferring the learned spatial reasoning capability to robotic control.In summary, this work makes the following contributions:


\begin{itemize}
    \item We propose a new \textbf{Spatial-Aware VLA Pretraining} paradigm that bridges 2D visual perception and 3D physical action through large-scale human video data.
    \item We construct \DataName, a dataset of human manipulation videos with 3D visual and action annotations enabling visual–physical alignment supervision.
    \item We present \ModelName, a dual-encoder VLA architecture pretrained with our paradigm, which significantly enhances spatial grounding and generalization in downstream robot tasks.
\end{itemize}

%% file: section/02_relatedwork.tex
\section{Related Work}

\subsection{VLA models}
Vision-Language-Action (VLA) models aim to leverage joint representations of visual observations and language instructions to enable action execution in physical environments\citep{brohan2022rt1,Chen2023PaLIXOS,li2024cogact,Zawalski24-ecot,ye2025dex1b,beingbeyond2025beingh0,goyal2024rvt2,he2025dexvlg,huang2025efficient,wan2023unidexgrasp++,deng2025graspvla,zhong2025dexgraspvla}.
Some prior works tokenize robot actions and finetune pretrained VLMs on robot datasets\citep{brohan2023rt2,kim2024openvla}. To support this pretraining paradigm, large-scale robot datasets have been introduced\citep{o2024open,bu2025agibot,khazatsky2024droid}. Other works incorporate additional action experts to improve the quality of action generation\citep{octo_2023,liu2024rdt,black2410pi0,bjorck2025gr00t}. Some approaches leverage demonstration videos for representation learning\citep{nair2022r3m,wang2024hpt,radosavovic2023real} or future state prediction to enhance action learning\citep{wu2024unleashing,bu2025agibot,Zhang2025DreamVLAAV}. More recently, models such as GR00T-N1\citep{bjorck2025gr00t} and $\pi_{0.5}$\citep{Intelligence202505AV} combine massive multimodal internet data with robot datasets for large-scale VLA pretraining, significantly improving generalization. In contrast, our work focuses on understanding the 3D physical space to better ground actions in environments with Spatial-Aware VLA Pretraining from human videos.

\subsection{3D Aware Models}
Recent advances in 3D vision-language models (3D VLMs) aim to extend the perceptual capacity of multimodal models beyond 2D images by incorporating 3D spatial information\citep{chen2023ll3da,zhu2024llava3d,spatialvlm2024chen,Fu2024SceneLLMEL,huang2024embodied,xu2025egodtm,3dllm}. Some approaches leverage datasets with explicit 3D annotations\citep{dai2017scannet,chen2020scanrefer,achlioptas2020referit_3d}, while others rely on geometric priors or depth estimation models to infer 3D structure from monocular visual inputs\cite{Bhat2023ZoeDepthZT,depth_anything_v1,dust3r_cvpr24,wang2025vggt,wang2025continuous}. These methods have improved the ability of VLMs to reason about spatial relationships, object geometry, and scene structure. However, the focus of existing 3D VLMs is primarily on perception rather than action. They enhance spatial understanding of static observations but do not explicitly establish the correspondence between 3D perception and the physical action space required for robot policy learning. Consequently, their utility in training VLA models remains limited.

Recent works have also explored enhancing VLAs with 3D spatial reasoning\citep{huang2024embodied}. For instance, 3D-VLA\citep{Zhen20243DVLAA3} introduces a generative world model with diffusion-based rendering of future images and point clouds, while SpatialVLA\citep{qu2025spatialvla} improves policy learning through ego-centric position encoding and adaptive action grids. 
Our work differs in that we aim to endow VLAs with 3D spatial understanding through a pretraining stage, by
leveraging 3D-aware annotations extracted from human videos and incorporating a 3D encoder to explicitly align 2D visual observations with 3D physical space. This Spatial-Aware Pretraining equips VLAs with stronger 3D grounding prior to downstream policy learning.



\subsection{Learning from human videos}
A line of research has explored leveraging human demonstration videos to facilitate robot policy learning. Some efforts adopt representation learning approaches to extract latent features from human activities\citep{nair2022r3m,bu2025agibot,xu2025egodtm}, yet such representations are typically implicit and provide limited guidance for explicit action grounding.
Other methods attempt to directly align the human action space with the robot action space\citep{kareer2024egomimic,yuan2025cross,singh2024hand,niu2025human2locoman,qiu2025humanoid}. However, these approaches suffer from embodiment mismatch, as the physical capabilities and kinematics of humans and robots differ significantly.
In addition, several works explore extracting interaction-related knowledge from human videos, such as affordances or grasping strategies\citep{chen2025vidbot,ma2025glover++,gavryushin2025maple}, which provide helpful cues but still stop short of explicitly bridging perception and action in 3D physical space.
Beyond these, Being-H0\citep{beingbeyond2025beingh0} also leverages human videos for VLA pretraining, but its primary focus lies in learning manipulation motion sequences.
In contrast, we highlight that, despite the embodiment gap, human videos contain rich information about how actions are executed in the 3D physical world across diverse visual contexts. 
Our work leverages this information as supervision for Spatial-Aware Pretraining, enabling VLA models to acquire grounded correspondences between 2D visual observations and 3D action space before downstream policy learning.

%% file: section/03-1_overview.tex
\section{Method}
\label{sec:method}

In this section, we first provide preliminaries on vision-language models (VLMs) and vision-language-action models (VLAs) (Sec.~\ref{sec:preliminary}).
Then we illustrate the \emph{Visual Phisical Alignment} in Sec.~\ref{sec:vipa}, including the spatial-aware dataset \DataName (Sec.~\ref{sec:dataset}), the model architecture \ModelName (Sec.~\ref{sec:model}) and the Spatial-Aware VLA Pretraining (Sec.~\ref{sec:pretrain}).
In Sec.~\ref{sec:post-method} we describe the post-training procedure for adapting the model to robot tasks.

\subsection{Preliminaries}
\label{sec:preliminary}
We begin by briefly formulating the setting of vision-language models (VLMs) and vision-language-action models (VLAs).  
A VLM is designed to map a set of visual inputs \(v = \{v_1, \dots, v_T\}\) (e.g., images or video frames) and a language instruction \(l\) (e.g., a question or command) into an aligned embedding space, producing a textual output \(y\) such as an answer or a caption:
\[
y = f_{\text{VLM}}(v, l).
\]

In contrast, a VLA extends this formulation to the action domain, where the model must not only capture the semantic alignment between vision and language but also generate executable actions in a 3D physical environment. Given visual observations \(v\) and an instruction \(l\), the VLA predicts an action chunk \(\mathbf{a}_t= \{\mathbf{a}_t^1, \dots, \mathbf{a}_t^H\}\):
\[
\mathbf{a}_t = f_{\text{VLA}}(v, l).
\]

This distinction highlights the central challenge addressed in this work: while VLMs excel at visual–semantic reasoning, VLAs additionally require grounding visual perception in 3D physical action space, which motivates our proposed Spatial-Aware VLA Pretraining paradigm.

\begin{figure}[t]
    \centering
    \includegraphics[width=0.85\linewidth]{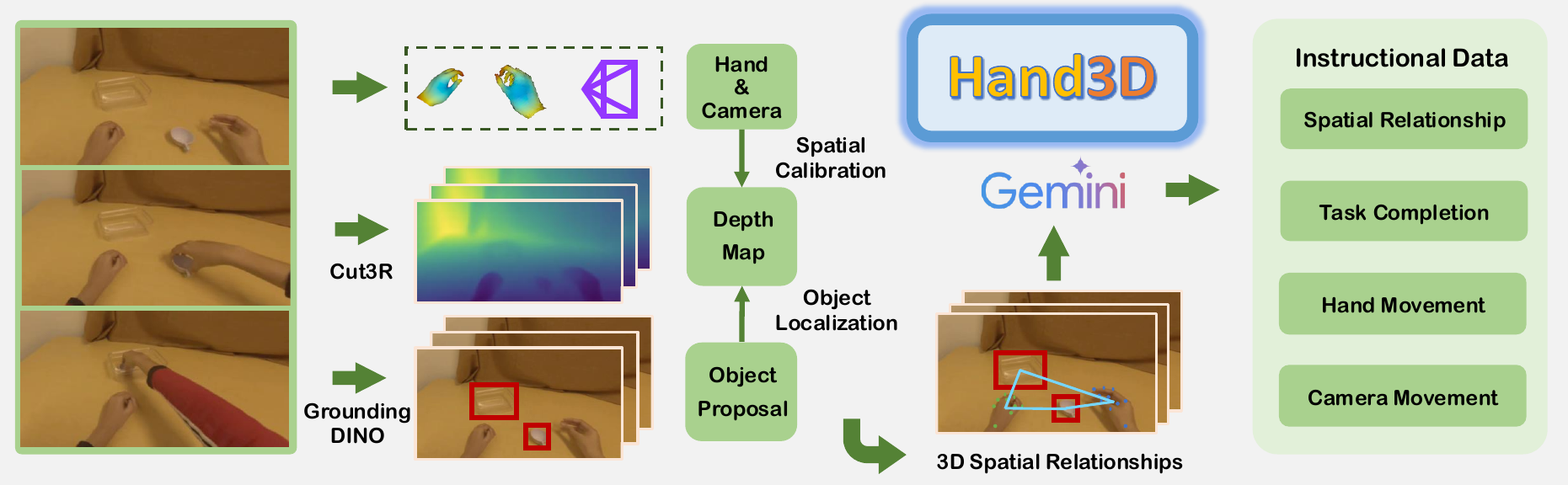}
    \caption{Overview of the \DataInstName. By integrating point cloud estimation, object localization, and hand pose annotations from human manipulation videos, we bridge 2D visual observations with 3D physical action space to provide visual-physical aligment supervision for VLA models.}
    \vspace{-3mm}
    \label{fig:hand3d}
\end{figure}

\subsection{Visual-Physical Alignment}
\label{sec:vipa}
\subsubsection{\DataName}
\label{sec:dataset}
Existing 3D VLMs typically obtain 3D supervision either from datasets with explicit 3D annotations or by estimating depth from 2D inputs. We follow their principles, but we focus on human demonstration videos, which inherently reveal how actions unfold in 3D space. These videos naturally provide rich source of visual-physical correspondences, allowing us to construct annotations that align 2D visual observations with action-relevant 3D information, forming the foundation for visual–physical alignment in VLA models.

\textbf{Human video collection.}
We start from the collection of human manipulation videos covering diverse tasks and interaction scenarios. Following UniHand\citep{beingbeyond2025beingh0}, we aggregate data from a wide range of different sources: \textbf{(1) Motion capture datasets}: Arctic\citep{fan2023arctic}, HOI4D\citep{liu2022hoi4d}, FPHA\citep{garcia2018first}, H2O\citep{kwon2021h2o}, OAKINK2\citep{zhan2024oakink2}, TACO\citep{liu2024taco}, Dex-YCB\citep{chao2021dexycb}, \textbf{(2) VR-recorded datasets}: EgoDex\citep{hoque2025egodex} and \textbf{(3) Pseudo-annotated datasets}: Taste-Rob\citep{zhao2025taste}, each providing recordings of humans performing object-centric manipulations and the corresponding hand motion sequences. To ensure consistency across heterogeneous datasets, we align all hand annotations to the unified MANO\citep{romero2017embodied} representation, which serves as a standard parametric model of human hand pose and shape.  For datasets with 3D hand joint annotations, we fit the MANO parameters via gradient optimization. For Taste-Rob, which is a video-only dataset, we estimate the hand MANO parameters with HaWoR\citep{zhang2025hawor}. This alignment enables us to obtain high-quality, normalized annotations of hand trajectories that can be directly associated with the corresponding visual frames.

\textbf{3D aware annotations.}
While prior works have explored learning from human videos, they largely remain confined to the 2D visual space, which restricts their ability to capture the physical grounding required for 3D action execution in robotic manipulation. We address this by extracting \textbf{3D visual annotations} and \textbf{3D action annotations} from human demonstrations.

As shown in Figure~\ref{fig:hand3d}, we construct \textbf{3D visual annotations} from human manipulation videos by combining point cloud estimation, object proposals, and hand pose information. Specifically, we first employ the Cut3R\citep{wang2025continuous} model to obtain a dense per-frame point cloud estimation $\mathcal{P}=\{(x_i, y_i, z_i)\}_{i=1}^N$, which provides 3D coordinates for each image pixel. Cut3R is chosen for its robustness in dynamic scenes and pretraining on human–object interaction data. For object localization, we apply Gemini-2.5-flash\citep{comanici2025gemini} to generate object proposals and GroundingDINO\citep{liu2023grounding} to obtain 2D bounding boxes $\mathcal{B}_o$, which approximate the spatial extent of each object in the image. By combining these bounding boxes with the depth estimation from $\mathcal{P}$, we localize each object in the 3D space and obtain its approximate spatial position.

In parallel, we utilize MANO-based hand pose annotations $\mathbf{m}=\{\theta, r, \tau, \beta \}$, provided in the dataset, together with camera extrinsics $[R \,|\, t]$ and intrinsics $K$. The MANO parameters are transformed into the camera coordinate system to yield 3D joint locations $\mathcal{J}_h=\{(x_j, y_j, z_j)\}_{j=1}^{21}$. These joint coordinates are projected onto the image plane,
$$
(u, v) = \Pi \!\left( K [R \,|\, t] (x,y,z)^\top \right),
$$
$$
\Pi(x',y',z')= \left(x'/z', \ y'/z' \right).
$$
By checking the visibility of the projected joints, we filter out frames where the hand is outside the view. Previous works have often relied on depth estimation models to introduce 3D information; however, the relative scale provided by such models typically does not match the actual physical space and may introduce misalignment. 
This mismatch can be problematic for action learning, since actions are executed at real-world physical scales. 
To unify scales between the relative point cloud estimation $\mathcal{P}$ and the physical space, we perform \textbf{scale calibration} by matching hand joint positions $\mathcal{J}_h$, which are absolute spatial positions, with corresponding point cloud coordinates. 

To formalize this process, let $\mathcal{J}_h^z = \{ j_k^z \}$ denote the set of hand joint depths in absolute physical space, and $\tilde{\mathcal{J}}_h^z = \{ \tilde{j}_k^z \}$ the corresponding depths obtained from the point cloud estimation $\mathcal{P}$. 
We estimate a scale factor $s$ as
\[
s = \operatorname{median}_{k \in \Omega} \left( j_k^z / \tilde{j}_k^z \right),
\]
where $\Omega$ is the set of valid joints determined by visibility and valid depth estimations. 
Applying $s$ to the point cloud $\mathcal{P}$ yields the calibrated representation $s\mathcal{P}$ which ensures that both hands and objects are represented in a consistent 3D physical coordinate system.
Consequently, we obtain spatially aligned annotations that encode the 3D geometry of hands and objects in consecutive frames.

\textbf{Instructional data curation.}
Building on the spatial annotations derived from human manipulation videos, we construct a set of vision–question–answering style labels that explicitly encode the relationship between 2D visual observations and 3D action information. This step is crucial for grounding visual inputs to physical actions, enabling VLA models to understand how 3D movements are manifested in visual space. To achieve this, we leverage Gemini-2.5-flash\citep{comanici2025gemini} to generate four complementary categories of question–answer pairs: (1) \textbf{Spatial Relationship:} Given information about a hand and an object within several consecutive frames, we generate natural language questions and answers that describe their 3D spatial relations in the last frame (e.g., relative position, distance, or contact state). (2) \textbf{Task Completion:} Conditioned on a task description and corresponding video frames, we formulate instructions about how the hand should move in 3D to interact with the target object. (3) \textbf{Hand Movement:} Given two frames, we annotate the hand’s 3D trajectory between them, including movement direction and distance. (4) \textbf{Camera Movement:} The relative 3D transformation of the camera for pairs of frames, expressed in terms of rotation direction and degree.

For the 3D spatial relationships or hand and camera movements used in Category 1--3 we represent each by the pair
\((\text{direction},\; \text{distance})\). Let \(\mathbf{v}= (x,y,z)\in\mathbb{R}^3\) denote the displacement vector that characterizes the relative 3D offset between two entities (e.g., object and hand, or hand positions across frames). The distance is defined as the Euclidean norm
$\mathrm{dist}(\mathbf{v}) = \lVert \mathbf{v} \rVert_2.$
The direction is derived from the unit vector \(\hat{\mathbf{v}} = \mathbf{v} / \lVert \mathbf{v}\rVert_2\) and discretized into axis-aligned language tokens with a component-wise threshold $\gamma$ to filter out negligible components:
\[
\begin{aligned}
\mathcal{D} = \{ \text{right/left if } |\hat{x}| > \gamma, \; \\
                \text{up/down if } |\hat{y}| > \gamma, \;\\
                \text{forward/backward if } |\hat{z}| > \gamma \},
\end{aligned}
\]

For camera movement, we consider the relative transformation between two camera poses $(R_1,t_1)$ and $(R_2,t_2)$, and decompose it into a rotation, represented by an axis-angle description $(\mathbf{u}, \phi)$, where $\mathbf{u}$ is the rotation axis and $\phi$ the rotation degree; and the translation, expressed as a direction vector and magnitude. 

\begin{minipage}[h]{0.50\textwidth}
Through this process, dense 3D geometry are transformed into compact, linguistically grounded labels directly used to produce VQA-style supervisory pairs, linking human visual demonstrations to physically grounded actions. Importantly, this formulation allows the model not only to perceive spatial configurations but also to reason about dynamic changes—hand motions, task-oriented manipulations, and even camera movements—thereby strengthening the bridge between 2D visual observations and 3D action policies.
With large-scale human data from nine heterogeneous human manipulation video sources, we uniformly sampled approximately 4K clips for annotation, ensuring coverage across diverse scenes, task types, and object interactions, and our pipeline produced around 300K instruction–answer pairs, 
which we denote as \DataInstName.
\end{minipage}
\hfill
\begin{minipage}[h]{0.45\textwidth}
\centering
\small
\captionof{table}{Data distribution of \DataInstName.\label{tab:supp-3d-visual}}
\vspace{-2mm}
\setlength{\tabcolsep}{8pt}
\scalebox{0.8}{
\begin{tabular}{lcc}
\toprule
\textbf{Category} & \textbf{Count} & \textbf{Proportion} \\
\midrule
\multicolumn{3}{l}{\textit{\textbf{Sources}}} \\
Arctic        & 100{,}772 & 33.5\% \\
OakInk2       & 45{,}926 & 15.3\% \\
TACO          & 39{,}087 & 13.0\% \\
H2O           & 32{,}390 & 10.8\% \\
HOI4D         & 28{,}977 & 9.6\% \\
EgoDex        & 28{,}200 & 9.4\% \\
FPHA          & 12{,}918 & 4.3\% \\
Taste-Rob     & 7{,}181 & 2.4\% \\
Dex-YCB       & 4{,}917 & 1.6\% \\
\midrule
\multicolumn{3}{l}{\textit{\textbf{Task Types}}} \\
Task Completion     & 206{,}409 & 68.7\% \\
Spatial Relations   & 74{,}887 & 24.9\% \\
Hand Movements      & 18{,}867 & 6.2\% \\
Camera Movements    & 205 & 0.1\% \\
\rowcolor{gray!20}{Total}         & 300{,}368 & 100\% \\
\bottomrule
\end{tabular}
}
\end{minipage}

\begin{figure}[h]
    \centering
    \includegraphics[width=0.85\linewidth]{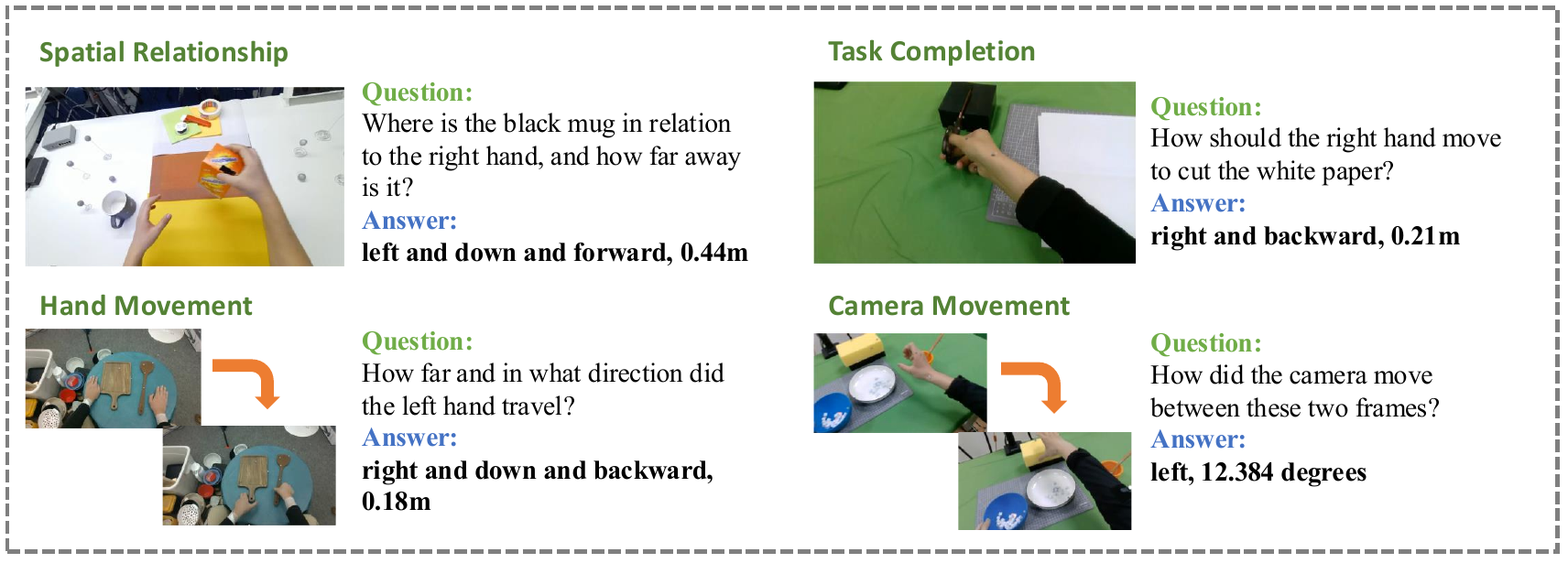}
    \vspace{-2mm}
    \caption{Examples of \DataInstName.}
    \label{fig:supp-example-data}
\end{figure}

We provide some examples of the \DataInstName in Figure~\ref{fig:supp-example-data}. The dataset covers four categories of manipulation-related tasks, encompassing diverse forms of spatial understanding essential for interaction. Such diversity ensures that the pretraining stage exposes the model to a wide range of manipulation scenarios, providing rich supervision for learning robust visual–physical alignment.

To complement the 3D visual annotations, we further construct \textbf{3D action annotations} that capture the dynamic aspect of human manipulation in physical space. From hand motion sequences in human videos, we extract wrist trajectories as $(x_t,y_t,z_t)$ spatial coordinate series, which are discretized via uniform binning into motion tokens $(m_1^x,m_1^y,m_1^z,...,m_t^x,m_t^y,m_t^z)$. Following UniHand\citep{beingbeyond2025beingh0}, we obtain text instructions for videos, yielding 4M video–instruction–motion pairs. Specifically, each video is first segmented into 10s chunks and Gemini-2.5-Flash\citep{comanici2025gemini} is applied to generate chunk-level and second-level annotations including task instructions and hand-object interactions. Then we utilize Gemini-2.5-Pro to construct three kinds of VQA-style tasks based on the annotations: instructional motion generation, motion translation and contextual motion prediction. We further filter out samples with insignificant 3D hand movement, resulting in a 1M curated dataset \DataTransName, providing fine-grained 3D motion patterns and spatial trajectories, as well as higher-level physical reasoning knowledge such as object affordance and task decomposition. The detailed composition and distribution of the \DataInstName and \DataTransName are provided in Table~\ref{tab:supp-3d-visual} and Table~\ref{tab:supp-3d-action}.

\begin{table}[ht]
\centering
\caption{Data distribution of \DataTransName.}
\small
\setlength{\tabcolsep}{7pt}
\vspace{-2mm}
\scalebox{0.8}{
\begin{tabular}{lcc}
\toprule
\textbf{Category} & \textbf{Count} & \textbf{Proportion} \\
\midrule
\multicolumn{3}{l}{\textit{\textbf{Sources}}} \\
EgoDex        & 758{,}050 & 73.5\% \\
Arctic        & 104{,}032 & 10.1\% \\
TACO          & 77{,}100 & 7.5\% \\
OakInk2       & 54{,}470 & 5.3\% \\
HOI4D         & 19{,}812 & 1.9\% \\
H2O           & 17{,}386 & 1.7\% \\
\midrule
\multicolumn{3}{l}{\textit{\textbf{Task Types}}} \\
Instructional Motion Generation     & 610{,}192 & 59.2\% \\
Contextual Motion Prediction   & 280{,}545 & 27.2\% \\
Motion Translation      & 140{,}113 & 13.6\% \\
\rowcolor{gray!20}{Total}         & 1{,}030{,}850 & 100\% \\
\bottomrule
\end{tabular}
}
\label{tab:supp-3d-action}
\end{table}

\subsubsection{\ModelName}
\label{sec:model}

\begin{figure}[t]
    \centering
    \includegraphics[width=0.75\linewidth]{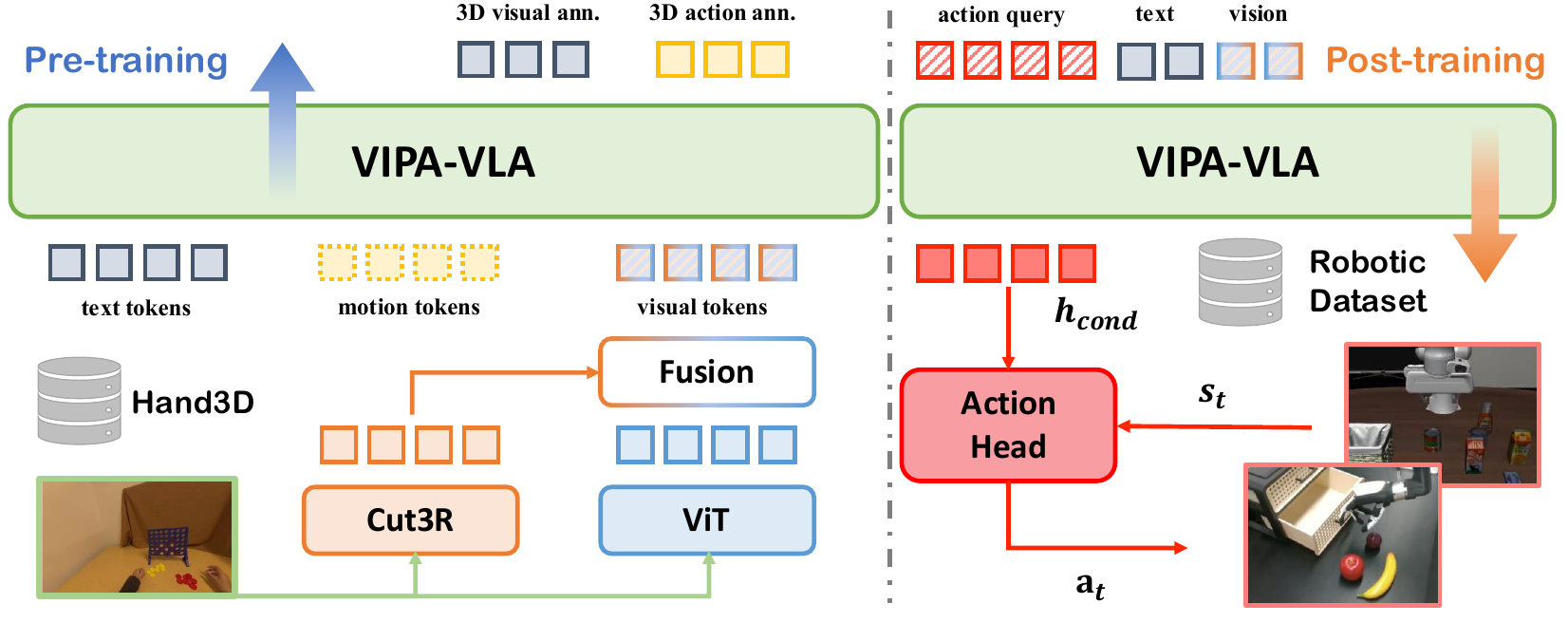}
    \caption{Model architecture of \ModelName. A dual-encoder including a semantic vision encoder and a 3D encoder produces fused spatial–semantic features through a cross-attention fusion layer. During pre-training, the vision tokens are aligned with text and motion tokens using 3D visual and 3D action annotations. During post-training, action queries interact with fused visual–language features to produce conditions, which is combined with the robot state and processed by a flow-matching action head to predict actions for robotic manipulation.}
    \vspace{-3mm}
    \label{fig:model-arch}
\end{figure}

Previous VLA models typically employ a semantic vision encoder to extract high-level visual semantics. However, such encoders lack the ability to provide features of 3D spatial structures. To address this limitation, we propose \ModelName, a dual-encoder architecture that integrates both semantic and spatial visual representations (Figure~\ref{fig:model-arch} left). Specifically, in addition to a semantic vision encoder, we incorporate a 3D vision encoder Cut3R \citep{wang2025continuous}, which provides explicit geometric understanding of the scene. The semantic encoder produces visual embeddings $\mathbf{V}_{sem}$, while the 3D encoder outputs spatial embeddings $\mathbf{V}_{spa}$. 

To effectively combine these complementary features, we introduce a fusion layer implemented as a cross-attention module. The fusion layer design is inspired by VLM-3R\citep{fan2025vlm3r}, which follows a cross-attention mechanism, where semantic visual features attend to 3D spatial features extracted by the 3D encoder. Given vision features $V_{sem}\in \mathbb{R}^{N_v\times d_v}$ and spatial features $V_{spa}\in \mathbb{R}^{N_s\times d_s}$, we first project them into a shared attention space, and then perform cross-attention where visual tokens query 3D spatial tokens. The output is projected back to the vision feature dimension using an output projection, resulting in $F_{spa}$. To integrate the spatial information while maintaining the pretrained visual semantics, the fusion layer applies a residual connection with a learnable scaling parameter $\alpha$:
$$
V_{f} = V_{sem} + \alpha F_{spa}
$$
Finally, dropout and layer normalization are applied to stabilize optimization.

To further enable the model to to understand fine-grained 3D motion trajectories extracted from human demonstrations, we extend the LLM’s token embedding space by introducing a set of motion tokens which discretize the 3D physical space in the second pretraining stage. Specifically, we tokenize each wrist trajectory point $(x_t,y_t,z_t)$ into three discrete motion tokens. To convert continuous 3D coordinates into discrete indices, we apply a uniform discretization over pre-defined bounded ranges. For each axis $a\in\{x,y,z\}$, we define a clipping range: $a\in[a_{min},a_{max}]$, and in our implementation, the $x$ and $y$ axes share the range $[-0.5,0.5]$, while the $z$ axis uses $[0,1]$. This effectively discretizes a $1m^3$ physical space directly in front of the camera into a structured token space. Each coordinate is discretized into one of $K$ bins, $m^a\in\{0,...,K-1\}$, where we use $K=1024$ in our experiments. Thus, each 3D waypoint $(x_t,y_t,z_t)$ is converted into a triplet of motion tokens $m_t^x,m_t^y,m_t^z$, yielding the final tokenized motion sequence $(m_1^x,m_1^y,m_1^z,...,m_t^x,m_t^y,m_t^z)$.

\subsubsection{Spatial-Aware VLA Pretraining}
\label{sec:pretrain}

To effectively align semantic perception with physical understanding, we design a \textbf{Spatial-Aware VLA Pretraining} strategy composed of two stages, enabling progressive learning of visual–physical alignment (Figure~\ref{fig:intro}). We first initialize our model using a pretrained VLM to inherit semantic understanding, while incorporating the pretrained 3D vision encoder and a randomly initialized fusion layer. In Stage 1, we freeze all pretrained parameters and train only the fusion layer using the 3D visual annotation VQA data. The objective is to align the semantic embeddings $\mathbf{V}_{sem}$ and spatial embeddings $\mathbf{V}_{spa}$ through the fusion layer, encouraging reasoning about 3D spatial relations.

In Stage 2, we extend the LLM vocabulary to include a set of motion tokens and then train the model on 3D action annotations. In this stage, the semantic and spatial encoders are frozen, and the LLM is trained to predict motion tokens conditioned on fused visual and textual inputs. This phase enables the model to acquire fine-grained spatial reasoning and action-level understanding, learning how visual cues correspond to physically grounded motion patterns. Through the Spatial-Aware VLA Pretraining, \ModelName progressively aligns 2D semantic perception, 3D spatial understanding, and action reasoning — achieving comprehensive visual–physical alignment for downstream robotic policy learning.

\begin{table*}[ht]
    \centering
    \caption{\textbf{Success rates (\%) on the LIBERO benchmark.} Results are reported for four task suites, with each suite evaluated across 500 trials. Experiments are conducted under both single-view and two-view input settings. All results are from public reports, except for GR00T N1.5$^*$, which we reproduce using the released model. The results of MaIL$^\dagger$ is from the ED-Ma version.}
    \small
    \setlength{\tabcolsep}{3pt}
    \vspace{-2mm}
    \begin{tabular}{l|c|ccccc}
    \toprule
        Model & Robo-PT & LIBERO-S (\%) & LIBERO-O (\%) & LIBERO-G (\%) & LIBERO-L (\%) & Avg. (\%) \\
        \midrule
        \multicolumn{6}{l}{\textcolor{gray}{\textit{Models with single-view input}}} \\
        TraceVLA\citep{zheng2025tracevla} & \color{red}{\Checkmark} & 84.6 & 85.2 & 75.1 & 54.1 & 74.8 \\
        OpenVLA\citep{kim2024openvla}& \color{red}{\Checkmark}   & 84.7 & 88.4 & 79.2 & 53.7 & 76.5 \\
        SpatialVLA\citep{qu2025spatialvla}& \color{red}{\Checkmark}   & 88.2 & 89.9 & 78.6 & 55.5 & 78.1 \\
        DiT Policy\citep{hou2024diffusion}& \color{red}{\Checkmark}  & 84.2 & 96.3 & 85.4 & 63.8 & 82.4 \\
        CoT-VLA\citep{Zhao2025CoTVLAVC}& \color{red}{\Checkmark}   & 87.5 & 91.6 & 87.6 & 69.0 & 83.9 \\
        ThinkAct\citep{huang2025thinkact}& \color{red}{\Checkmark}  & 88.3 & 91.4 & 87.1 & 70.9 & 84.4 \\
        TriVLA\citep{liu2025trivla}& \color{red}{\Checkmark}   & 91.2 & 93.8 & 89.8 & 73.2 & 87.0 \\
        4D-VLA\citep{zhang20254d}& \color{red}{\Checkmark}   & 88.9 & 95.2 & 90.9 & \underline{79.1} & 88.6 \\
        GR00T N1.5$^*$\citep{bjorck2025gr00t}& \color{red}{\Checkmark}  & \underline{91.4} & \textbf{97.6} & \underline{94.0} & \textbf{85.6} & \underline{92.1} \\
        \rowcolor{gray!20} \textbf{\ModelName}& \color{blue}{\XSolidBrush} & \textbf{92.6} & \underline{97.2} & \textbf{94.2} & \textbf{85.6} & \textbf{92.4} \\
        \midrule
        \multicolumn{6}{l}{\textcolor{gray}{\textit{Models with two-view input}}} \\ 
        MaIL$^\dagger$\citep{jia2024mail}& \color{blue}{\XSolidBrush}  & 74.3 & 90.1 & 81.8 & 78.6 & 83.5 \\
        $\pi_0$-FAST \citep{pertsch2025fast}& \color{red}{\Checkmark}  & 96.4 & 96.8 & 88.6 & 60.2 & 85.5 \\
        MolmoAct\citep{lee2025molmoact}& \color{red}{\Checkmark}  & 87.0 & 95.4 & 87.6 & 77.2 & 86.6 \\
        GR00T N1\citep{bjorck2025gr00t}& \color{red}{\Checkmark}  & 94.4 & 97.6 & 93.0 & 90.6 & 93.9 \\
        $\pi_0$ \citep{black2410pi0}& \color{red}{\Checkmark}  & \underline{98.0} & 96.8 & 94.4 & 88.4 & 94.4 \\
        UniVLA\citep{wang2025unified}& \color{red}{\Checkmark}  & 95.4 & \textbf{98.8} & 93.6 & 94.0 & 95.5 \\
        $\pi_{0.5}$\citep{black2025pi_}& \color{red}{\Checkmark}  & \textbf{98.8} & 98.2 & \textbf{98.0} & \underline{92.4} & \textbf{96.9} \\
        \rowcolor{gray!20} \textbf{\ModelName}& \color{blue}{\XSolidBrush}  & 96.6 & \underline{98.6} & \underline{97.0} & \textbf{95.0} & \underline{96.8} \\
    \bottomrule
    \end{tabular}
    \label{tab:exp-libero}
\end{table*}

\subsection{Post-Training}
\label{sec:post-method}
After pretraining, \ModelName has acquired alignment between 2D observations and 3D physical/action representations. In the post-training stage, we fine-tune this initialization on downstream robot tasks by attaching an \emph{action head} and training it to produce executable action chunks. For implementation, we use a diffusion transformer (DiT) model, parameterized with $\theta$.

Let \(v\) and \(l\) denote visual frames and language instruction, and let \(\mathcal{Q}_a\) be a fixed set of action queries. We extract conditional context from the pretrained VLM backbone as
$h_{\mathrm{cond}} = \mathrm{VLM}_\phi(v, l, \mathcal{Q}_a),$
where the hidden states corresponding to \(\mathcal{Q}_a\) are used as the DiT conditioning. The ground-truth action chunk for timestep \(t\) is denoted by \(\mathbf{a}_t\), and the robot state embedding by \(s_t\). Following flow matching\citep{lipman2023flow}, we generate a noisy action trajectory by linearly interpolating between a random noise vector \(\boldsymbol{\epsilon}\) and the ground-truth action:
\[
\tilde{\mathbf{a}}_t^{(\tau)} \;=\; (1-\tau)\,\boldsymbol{\epsilon} \;+\; \tau\,\mathbf{a}_t, 
\quad \tau \sim \mathcal{U}(0,1).
\]
The DiT receives the concatenation of \(\tilde{\mathbf{a}}_t^{(\tau)}\) and the state embedding \(s_t\) as its input representation:
$h_{\mathrm{DiT}} = \operatorname{concat}\!\big(\tilde{\mathbf{a}}_t^{(\tau)},\, s_t\big).$ Conditioned on \(h_{\mathrm{cond}}\), the DiT predicts the instantaneous flow vector \(\mathbf{v}_\theta\). The training objective minimizes the deviation between this prediction and the oracle transport direction given by \(\mathbf{a}_t - \boldsymbol{\epsilon}\):
\[
\mathcal{L}_{\mathrm{FM}} \;=\; \mathbb{E}_{\mathbf{a}_t,\tau,\boldsymbol{\epsilon},v,l}
\left[ \left\| 
\mathbf{v}_\theta - (\mathbf{a}_t - \boldsymbol{\epsilon})
\right\|_2^2 \right].
\]
During training, only the LLM backbone and the action head \(f_\theta\) are updated to adapt the model to action execution. The illustration is shown in Figure~\ref{fig:model-arch} (right).

%% file: section/05_experiment.tex
\section{Experiment}

\subsection{Implementation Details}
\ModelName is initialized from InternVL3.5-2B\citep{wang2025internvl3_5}. 
During the first pre-training stage, for the \textit{spatial relationship} and \textit{task completion} tasks, the model is given 1--4 randomly sampled consecutive frames as input, while for the \textit{hand movement} and \textit{camera movement} tasks, two frames are provided. All video frames are sampled at 1 fps. In the second pre-training stage and the post-training stage, we use single frame as visual input.

The training consists of three stages, and the training detials are as follows:

\paragraph{Stage 1} We pretrain \ModelName with \DataInstName for a single epoch in the first stage, cost around 6 hours on 8 NVIDIA A800 GPUs. Only the fusion layer is optimized, with AdamW optimizer using the learning rate of $1e-5$. We set warm up ratio to $0.03$ and the weight decay to $0.01$, with a cosine scheduler. The global batch size is 32. For the fusion layer, the spatial scaling parameter $\alpha$ is initialized as $0.5$. The image frames are resized to $448 \times 448$.

\paragraph{Stage 2} In the second stage we use \DataTransName to optimize the fusion layer and the LLM backbone of \ModelName, requiring around 20 hours on 8 NVIDIA A800 GPUs. The training hyper-parameters are kept the same as in the first stage.

\paragraph{Stage 3} In the post-training stage, we freeze the visual encoder and the 3D encoder, and the global batch size is $128$ (for LIBERO and real robot tasks) or $256$ (for RoboCasa), with a learning rate $5e-5$. The warm up ratio is $0.05$ and the weight decay is $1e-5$. For both the LIBERO benchmark and the real robot tasks, we train \ModelName for 30K steps, requiring around 5 hours on 8 NVIDIA A800 GPUs. For the RoboCasa benchmark, we train for 60K steps, taking approximately 40 hours on 8 NVIDIA A800 GPUs.

\subsection{Simulation Robot Tasks}
\paragraph{Benchmarks.}
We first evaluate \ModelName's performance for robotic manipulation in simulation environments, which assesses how well the model can generalize spatially grounded reasoning from human demonstrations to robotic scenarios. Specifically, we adopt \textbf{LIBERO}\citep{liu2023libero}, which is a standard benchmark for evaluating robot manipulation capabilities, emphasizing robustness and generalization, and \textbf{RoboCasa}\citep{nasiriany2024robocasa}, which poses significantly greater challenges than LIBERO—featuring more diverse layouts, cluttered scenes, and more complex visual observations, demanding more accurate and robust 3D spatial understanding from VLA models. \textbf{LIBERO} consists of four task suites: \textit{Spatial}, \textit{Object}, \textit{Goal} and \textit{Long}, covering different task settings and scene layouts.

\paragraph{Results on LIBERO}
We report the average success rates across 500 trials for each task suite in Table~\ref{tab:exp-libero}. \ModelName achieves strong performance across all LIBERO task suites under both single-view and two-view input settings. This demonstrates that the Spatial-Aware VLA Pretraining learned from human videos generalizes effectively to simulated robotic environments.
Notably, our model does not rely on any robot data during pretraining, yet it surpasses a number of previous methods and achieves results comparable to $\pi$-series and GR00T models, which are strong baselines pretrained on large-scale robot datasets. This indicates that the spatial reasoning ability acquired from human demonstrations—bridging 2D visual cues with 3D spatial understanding—can significantly help robotic manipulation even in simulation. Among baselines that explicitly model spatial reasoning, such as SpatialVLA\citep{qu2025spatialvla}, TraceVLA\citep{zheng2025tracevla}, and MolmoAct\citep{lee2025molmoact}, our approach consistently performs better, highlighting the effectiveness of learning visual–physical alignment from diverse human demonstrations.

\paragraph{Results on RoboCasa}
For experiments on RoboCasa, we adopt a three-view input configuration and train \ModelName using only 50 human demonstrations per task. As shown in Table~\ref{tab:supp-robocasa}, across the diverse RoboCasa tasks, \ModelName achieves the best overall performance, demonstrating strong generalization under multi-view observations and limited demonstration data. Notably, our model attains significant improvements in the \textit{Doors/Drawers} categories ($+9.9\%$), which require precise spatial localization. These results highlight that the Spatial-Aware Pretraining effectively equips the model with more reliable 2D–3D grounding, enabling accurate spatial reasoning even in visually diverse and geometrically challenging environments. 

\begin{table}[t]
    \centering
    \vspace{-2mm}
    \caption{\textbf{Success rates (\%) on the RoboCasa benchmark.} Models are evaluated on 24 tasks (8 for \textit{Pick \& Place}, 6 for \textit{Doors / Drawers}, 10 for \textit{Others}), with each task evaluated across 50 trials.}
    \small
    \setlength{\tabcolsep}{3pt}
    \vspace{-2mm}
    \begin{tabular}{l|cccc}
    \toprule
        Model & Pick \& Place & Doors / Drawers & Others & Avg. \\
        \midrule
        GR00T N1 & 18.6 & 50.2 & 39.1 & 36.0 \\
        $\pi_{0.5}$ & \textbf{21.5} & 57.8 & 44.9 & 41.4 \\
        \rowcolor{gray!20} \textbf{\ModelName}  & 20.8 & \textbf{67.7} & \textbf{52.8} & \textbf{45.8} \\
    \bottomrule
    \end{tabular}
    \vspace{-2mm}
    \label{tab:supp-robocasa}
\end{table}

\subsection{Real Robot Tasks}
\label{sec:exp-real}
We further evaluate our model in real-world robotic manipulation tasks, which provide a more realistic 3D physical setting and require more accurate spatial perception and action grounding, making them ideal for assessing the benefits of our Spatial-Aware VLA Pretraining with human videos. The experiments are conducted with a 7-DoF Franka Research 3 arm, a 6-DoF Inspire hand, and two RealSense L515 cameras for visual observation.

We design three manipulation tasks to test different aspects of spatial perception and reasoning: (1) \textbf{Put-Three-Obj} – sequentially place three fruits into a drawer, assessing the ability to localize and manipulate multiple objects; (2) \textbf{Wipe-Board} – use a cloth to remove all pen marks from a whiteboard, requiring flexible spatial reasoning over irregularly shaped target regions; (3) \textbf{Water-Plant} – pick up a watering can and water a plant, demanding precise spatial localization and control of fine-grained 3D motion. Each task is further decomposed into several sub-tasks, which are used to measure fine-grained execution performance. And we also design unseen environment setups for each task for evaluation of the generalization abilities. Details of the configurations are as follows:
\begin{itemize}
    \item \textbf{Put-Three-Obj}: The task requires the robot to open the drawer, sequentially pick and place three fruits (apple, banana, plum) from the table into the drawer, and close the drawer. We define five sub-tasks: \emph{open drawer}, \emph{pick \& place apple}, \emph{pick \& place banana}, \emph{pick \& place plum}, and \emph{close drawer}. For the unseen environment setup, the table surface is modified by covering it with tablecloths of unseen colors.
    \item \textbf{Wipe-Board}: The task requires the robot to picks up a cloth and wipes the pen marks off the whiteboard. There are three sub-tasks defined: \emph{pick up cloth}, \emph{wipe pen marks}, and \emph{clean entire whiteboard}. For the unseen environment setup, we change the color of the marker used to draw the pen marks.
    \item \textbf{Water-Plant}: The task requires the robot to pick up a watering can, locate the plant, and water it by pressing the spray handle. There are three sub-tasks defined: \emph{pick up watering can}, \emph{locate plant}, and \emph{press spray handle}.
\end{itemize}

\begin{wrapfigure}{r}{0.5\textwidth}
    \centering
    \includegraphics[width=0.95\linewidth]{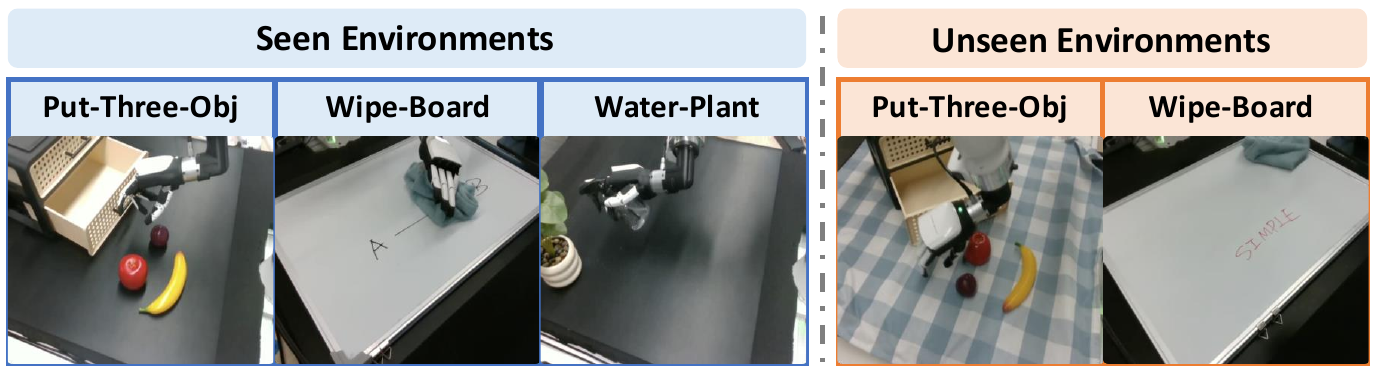}
    \vspace{-3mm}
    \caption{Real Robot Task Settings.}
    \label{fig:exp:real-settings}
\end{wrapfigure}
The illustration of the tasks are shown in Figure~\ref{fig:exp:real-settings}.
For each task, we collect 50 teleoperated trajectories for post-training and conduct 10 evaluation trials to measure task success rates. During evaluation, we randomize the placement of objects on the table and the distribution of pen marks on the whiteboard for each trial. This setup ensures that the learned policies are tested on their generalization ability rather than memorization of fixed configurations. Besides the success rates of the whole tasks, we also report the success rates computed by the proportion of completed sub-tasks, averaged across all trials.


As shown in Table~\ref{tab:real-robot-exp}, \ModelName achieves the best overall performance across all real-world manipulation tasks, demonstrating that the spatial understanding acquired during pretraining effectively helps real robotic manipulations, leading to more robust and generalizable behavior in real-world manipulation scenarios. Compared with the ablation baseline InternVL3.5, our model achieves substantial improvements on all tasks, confirming the effectiveness of our proposed Spatial-Aware Pretraining and dual-encoder architecture in enhancing spatial grounding and generalization. 
Although \ModelName achieves lower complete-task success rate on Put-Three-Obj, it attains higher sub-task success rate. This is because baseline models often fail early in the sequence—sometimes unable to complete even the first step—indicating less stable spatial reasoning and control. In contrast, \ModelName exhibits more consistent progress across task stages, reflecting stronger robustness in real-world execution.

\begin{table}[h]
\centering
\caption{Success rates (\%) of \ModelName v.s. baseline models on real robot tasks. Results are shown as sub-task / whole-task success rates.}
\small
\vspace{-2mm}
\setlength{\tabcolsep}{6pt}
\scalebox{0.9}{
\begin{tabular}{l|ccc}
\toprule
{Task} & {Put-Three-Obj} & {Wipe-Board} & {Water-Plant}  \\
\midrule
GR00T N1.5~\cite{bjorck2025gr00t} & 48\% / \textbf{40\%} & 57\% / 30\%  & 53\% / 30\% \\
Being-H0\citep{beingbeyond2025beingh0} & 38\% / 20\% & 40\% / 10\% & 37\% / 20\%\\
InternVL3.5~\cite{wang2025internvl3_5} & 34\% / 10\% & 43\% / 10\% & 37\% / 20\% \\
\rowcolor{gray!20} \textbf{\ModelName} & \textbf{52\%} / 10\% & \textbf{83\%} / \textbf{60\%} & \textbf{57\%} / \textbf{50\%} \\
\bottomrule
\end{tabular}
\vspace{-2mm}
\label{tab:real-robot-exp}
}
\end{table}

\begin{table}[h]
\centering
\caption{Success rates (\%) of \ModelName v.s. baseline models on real robot tasks with unseen environments. Results are shown as sub-task / whole-task success rates.}
\small
\vspace{-2mm}
\setlength{\tabcolsep}{6pt}
\scalebox{0.9}{
\begin{tabular}{l|cc}
\toprule
{Task} & {Put-Three-Obj-Unseen} & {Wipe-Board-Unseen}  \\
\midrule
GR00T N1.5~\cite{bjorck2025gr00t} & 28\% / 0\% & 43\% / 10\% \\
Being-H0\citep{beingbeyond2025beingh0} & 16\% / 0\% & 33\% / 10\% \\
InternVL3.5~\cite{wang2025internvl3_5} & 42\% / 10\% & 40\% / 10\%  \\
\rowcolor{gray!20} \textbf{\ModelName} & \textbf{44\%} / \textbf{20\%} & \textbf{83\%} / \textbf{50\%}  \\
\bottomrule
\end{tabular}
\vspace{-2mm}
\label{tab:real-robot-exp-unseen}
}
\end{table}

We further evaluate the models in unseen environments. As shown in Table~\ref{tab:real-robot-exp-unseen}, \ModelName maintains strong performance when evaluated in unseen visual environments, significantly outperforming all baselines. While other models often exhibit a sharp degradation in performance—often failing to complete even the initial steps—\ModelName retains high sub-task and whole-task success rates, demonstrating enhanced robustness and generalization. These results highlight that the Spatial-Aware Pretraining effectively equips the model with transferable 3D spatial understanding learned from human videos. By grounding 2D visual observations into 3D physical representations, our model generalizes more reliably to new scenes.

\begin{figure}[h]
    \centering
    \includegraphics[width=0.9\linewidth]{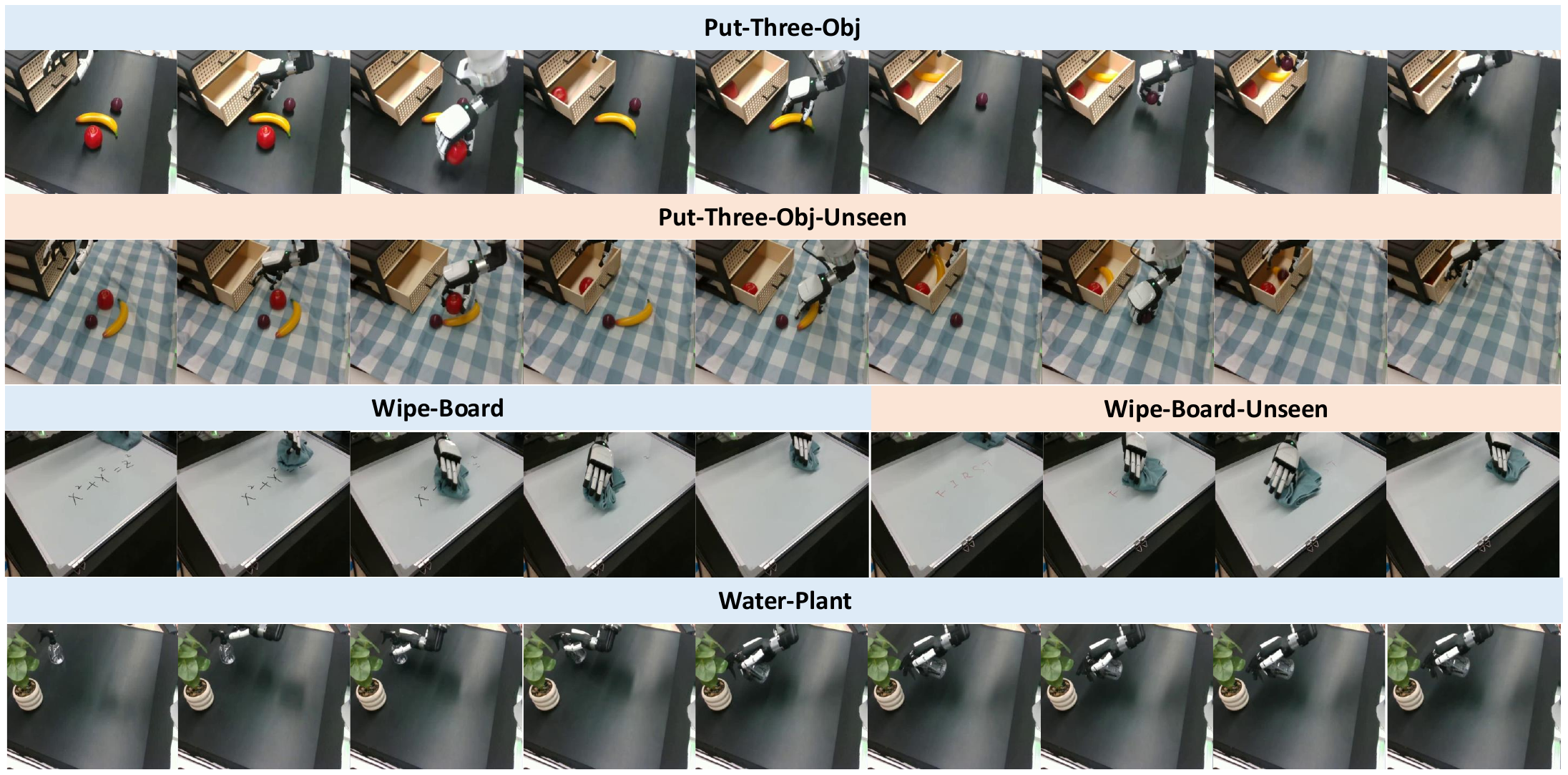}
    \vspace{-2mm}
    \caption{Qualitative examples of \ModelName performing real robot tasks.}
    \label{fig:supp-real-case}
\end{figure}

We show qualitative examples of \ModelName executing real robot tasks in Figure~\ref{fig:supp-real-case}. Across diverse scenes and object configurations, the model demonstrates reliable spatial grounding and robust generalization: it accurately localizes objects, adjusts to varying layouts, and completes multi-step manipulation sequences with consistent trajectories. These examples highlight the model’s ability to transfer its spatial-aware pretraining to real-world settings, enabling precise and stable task execution even under unseen spatial variations.

\begin{figure}[h]
    \centering
    \includegraphics[width=0.85\linewidth]{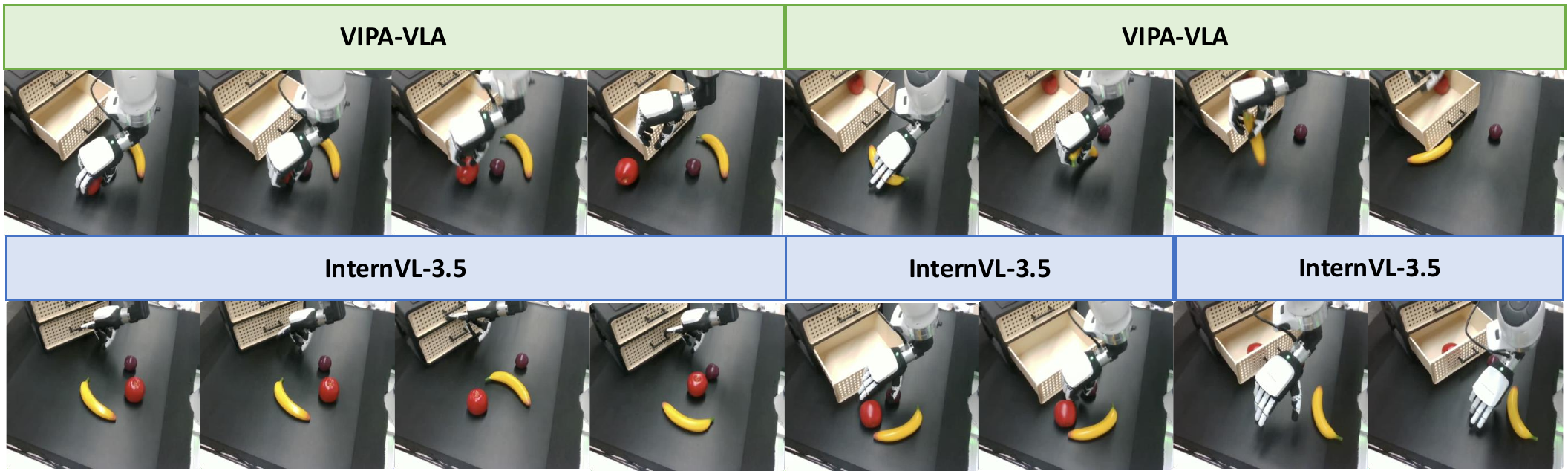}
    \vspace{-2mm}
    \caption{Failure examples of \ModelName and InternVL-3.5 performing real robot tasks.}
    \label{fig:supp-fail-case}
\end{figure}

We also provide analysis of typical failure cases of \ModelName and the baseline model InternVL-3.5 in Figure~\ref{fig:supp-fail-case}. For \ModelName, failures generally occur in fine-grained manipulation steps—such as slight misalignment when grasping a small object—while the overall spatial localization and action planning remain correct. These errors often arise from subtle control inaccuracies rather than misunderstanding the physical space.
In contrast, the baseline model frequently fail due to inaccurate spatial grounding: they may reach toward incorrect regions or misjudge object positions that reflect poor 2D–3D correspondence. Such failures indicate weaker spatial understanding, causing the policy to be stuck before meaningful manipulation begins.

\subsection{Ablation Studies}
\paragraph{Architecture and the Spatial-Aware VLA Pretraining.}
We conduct ablation studies on the \textbf{LIBERO} benchmark to evaluate the effectiveness of our model architecture and the proposed Spatial-Aware VLA Pretraining. This analysis isolates the contributions of the dual-encoder design and the pretraining strategy with visual-physical alignment.

\begin{table}[h]
    \centering
    \vspace{-2mm}
    \caption{Ablations (\%) on the LIBERO benchmark.}
    \small
    \setlength{\tabcolsep}{3pt}
    \vspace{-2mm}
    \begin{tabular}{l|ccccc}
    \toprule
        Model & Spa. & Obj. & Goal & Long & Avg. \\
        \midrule
        \rowcolor{gray!20} \textbf{\ModelName}  & \textbf{92.6} & \textbf{97.2} & \textbf{94.2} & \textbf{85.6} & \textbf{92.4} \\
        \midrule
        – Pretraining & 90.8  & 97.0 & 93.0 & 84.0 & 91.2 {\color{red!60!black}\scriptsize{(-1.2\%)}} \\
        – Dual Encoder & 90.0 & \textbf{97.2} & 92.4 & 81.8 & 90.4 {\color{red!60!black}\scriptsize{(-2.0\%)}}\\
        – Both & 89.2 & 95.2 & 90.0 & 80.4 & 88.7 {\color{red!60!black}\scriptsize{(-3.7\%)}} \\
    \bottomrule
    \end{tabular}
    \vspace{-2mm}
    \label{tab:exp-ablation}
\end{table}

As shown in Table \ref{tab:exp-ablation}, both the Spatial-Aware Pretraining on human videos and the proposed dual-encoder architecture contribute to overall performance improvements. Removing either component leads to a consistent performance drop across all task suites. The pretraining provides the model with rich 3D motion and perception priors learned from human demonstrations, while the dual encoder enables more effective fusion of 3D visual representations. When combined, these two designs further enhance the model’s spatial reasoning and generalization, resulting in the strongest performance.

\paragraph{Effect of Spatial-Aware VLA Pretraining.}
Here we first evaluate the effectiveness of the first stage pretraining on \textbf{3D visual annotations} in improving spatial understanding capabilities. Specifically, we assess how well the model can reason about spatial relations and 3D geometry in visual scenes. We report results on the \DataTestName, which consists of 2K VQA pairs from unseen videos not included in the training corpus. The evaluation focuses on two aspects: \textit{distance accuracy} and \textit{direction accuracy}. 
Direction accuracy is measured by assessing the correctness of predictions along the three principal axes; each axis contributes a binary score of 0 or 1. 
Distance accuracy is quantified by the average prediction error in meters. 

We compare \ModelName-PT against two ablations: 
(1)\textbf{InternVL3.5}: the backbone VLM without additional pretraining.
(2)\textbf{InternVL3.5}+\DataName: The backbone model pretrained with \DataName, without the proposed dual encoder architecture.
This setup allows us to disentangle the contribution of \DataName pretraining and the architecture of \ModelName. For InternVL3.5, we provide an example answer in the prompt for valid answering format.

\begin{table}[h]
\centering
\small
\vspace{-1mm}
\caption{Evaluation of 3D spatial understanding.}
\vspace{-2mm}
\begin{tabular}{lcc}
\toprule
\textbf{Model} & \textbf{Dist. Err. (m) $\downarrow$} & \textbf{Dir. Scr. $\uparrow$} \\
\midrule
InternVL3.5            & 0.18  & 1.22/3 \\
{InternVL3.5 +\DataName} & 0.14 & 1.75/3 \\
\midrule
\rowcolor{gray!20} \ModelName-PT           & \textbf{0.12} & \textbf{1.82/3} \\
\bottomrule
\end{tabular}
\vspace{-2mm}
\label{tab:3d_spatial_results}
\end{table}

\begin{figure}[h]
    \centering
    \includegraphics[width=0.65\linewidth]{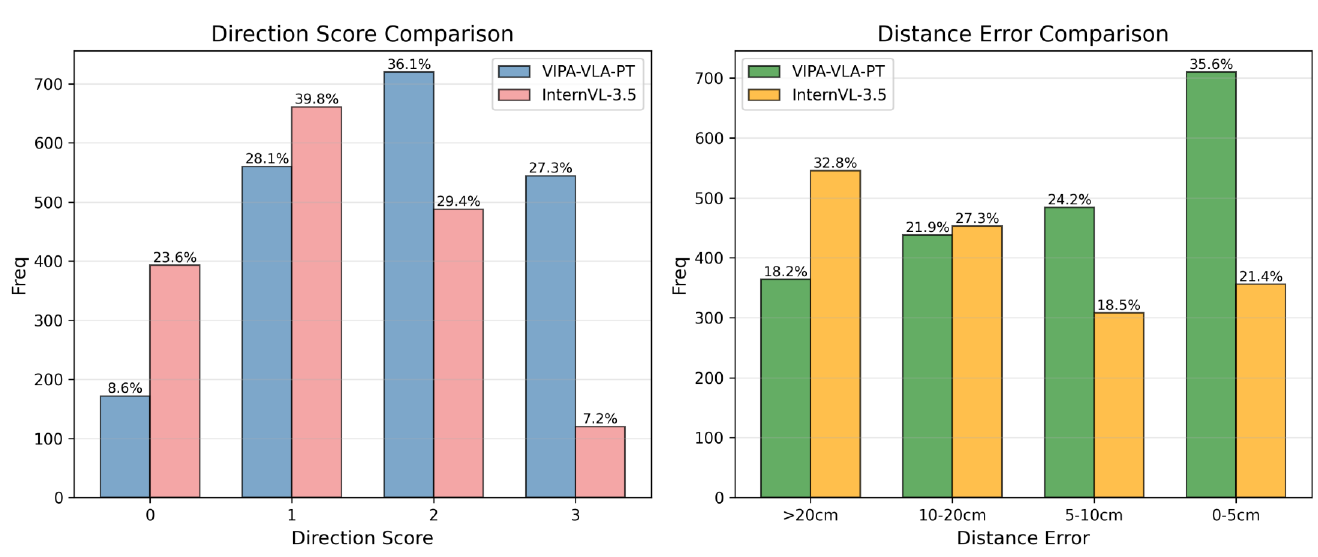}
    \caption{Comparison between \ModelName-PT and InternVL3.5 on \DataTestName.
    Left: histogram of direction scores;  
    Right: histogram of distance errors}
    \vspace{-3mm}
    \label{fig:exp:3d-pt-ablation}
\end{figure}

From the results shown in Table~\ref{tab:3d_spatial_results}, the backbone InternVL3.5 performs poorly on the spatial reasoning tasks, indicating its limited understanding of spatial relationships. Pretraining on \DataName notably improves performance, confirming that 3D visual annotations effectively enhance the model’s spatial grounding through visual-physical alignment. Our proposed \ModelName further achieves a consistent performance gain, demonstrating that introducing the 3D visual encoder and the fusion mechanism brings additional benefits beyond data-level supervision. Notably, only the fusion layer is trained at this stage, showing that aligning semantic and spatial visual features alone already yields substantial gains.

We further show detailed comparison between \ModelName-PT and InternVL3.5 in Figure~\ref{fig:exp:3d-pt-ablation}. After Spatial-Aware Pretraining on \textbf{3D visual annotations}, our model makes substantially fewer severe errors and attains highly accurate predictions in about 30\% of the cases, reflecting precise and reliable 3D spatial grounding ability obtained in this stage.

\begin{figure}[h]
    \centering
    \includegraphics[width=0.85\linewidth]{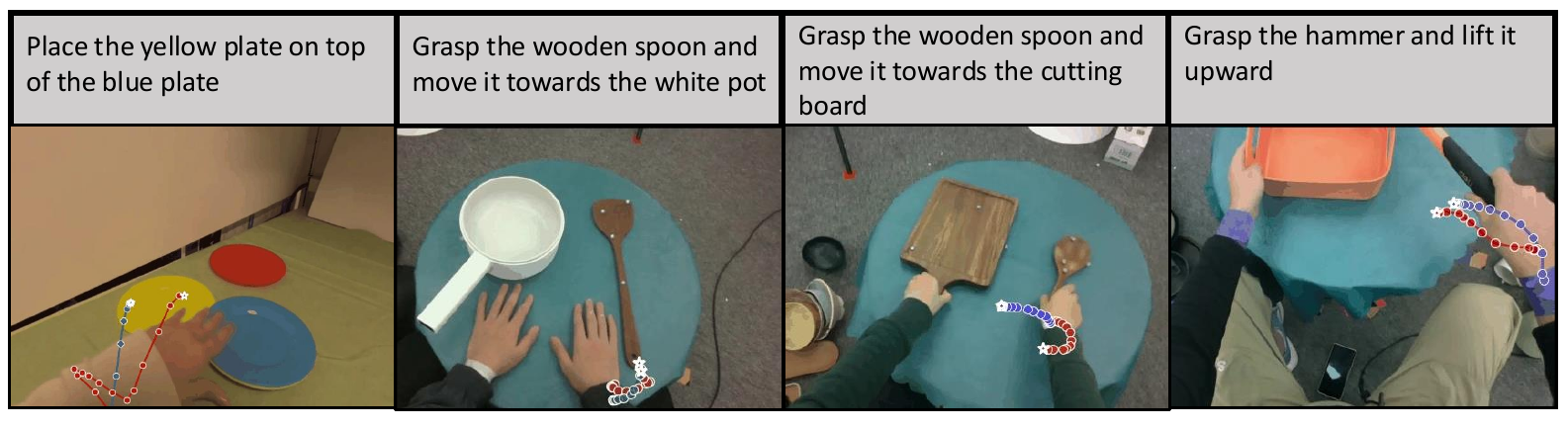}
    \vspace{-2mm}
    \caption{Visualization of the predicted motion trajectories from \ModelName after second stage pretraining (blue lines) and the ground-truth trajectories (red lines).}
    \label{fig:exp:3d-trans-pred}
\end{figure}

We then show qualitative results of the second stage pretraining on \textbf{3D action annotations} in Figure~\ref{fig:exp:3d-trans-pred}. The predicted trajectories closely align with the instructions, showing that the model effectively understands the mapping between visual observations and physical space. Compared to the ground-truth trajectories, which often exhibit noisy and redundant motion due to natural human variability, our model produces smoother and more goal-directed trajectories. Moreover, the predictions reflect learned affordance knowledge — for example, when manipulating a wooden spoon, the model grasps it near the handle end.

%% file: section/06_conclusion.tex
\section{Conclusion}

In this work, we address the critical 2D–3D gap in vision-language-action (VLA) models through \textbf{Spatial-Aware VLA Pretraining} from large-scale human videos. We introduce \DataName, a dataset providing 3D visual and action annotations derived from human manipulation, and propose \ModelName, a dual-encoder architecture designed for \emph{visual-physical alignment}. The proposed paradigm effectively aligns 2D perception with 3D physical understanding, yielding strong improvements in spatial reasoning and downstream robotic manipulation.
Our results validate the effectiveness of Spatial-Aware Pretraining from human demonstrations. In future, this pretraining paradigm can be combined with robot data pretraining to achieve a more comprehensive and effective overall pretraining strategy.